\documentclass[runningheads]{llncs}
\usepackage{graphicx}
\usepackage{orcidlink}
\usepackage{tikz}
\usepackage{comment}
\usepackage{amsmath,amssymb} 
\usepackage{color}
\usepackage{booktabs}
\usepackage{multirow}
\usepackage{float}
\usepackage{hyperref}

\usepackage[accsupp]{axessibility}  

\usepackage{overpic}
\usepackage{enumitem} 
\usepackage{overpic} 
\usepackage{xcolor}

\newcommand{\etal}{\textit{et al.}}
\newcommand{\ie}{\textit{i.e.}}

\begin{document}
\pagestyle{headings}
\mainmatter
\def\ECCVSubNumber{4401}  

\title{Neural Color Operators \\for Sequential Image Retouching} 

\titlerunning{ECCV-22 submission ID \ECCVSubNumber} 
\authorrunning{ECCV-22 submission ID \ECCVSubNumber} 
\author{Anonymous ECCV submission}
\institute{Paper ID \ECCVSubNumber}

\titlerunning{Neural Color Operators}
\author{Yili Wang\inst{1}\orcidlink{0000-0003-1364-0480}\thanks{Work done during Yili Wang's internship at VIS, Baidu.} \and
Xin Li\inst{2} \and
Kun Xu\inst{1}\orcidlink{0000-0002-2671-4170}\thanks{Kun Xu is the corresponding author.} \and
Dongliang He\inst{2}\orcidlink{0000-0002-1129-8610} \and
Qi Zhang\inst{2} \and
\\ Fu Li\inst{2} \and
Errui Ding\inst{2}\orcidlink{0000-0002-1867-5378}}
\authorrunning{Yili Wang et al.}
\institute{BNRist, Department of CS\&T, Tsinghua University \\ \and
Department of Computer Vision Technology (VIS), Baidu Inc.\\ \email{yiliw.thu@gmail.com, lixin41@baidu.com,  xukun@tsinghua.edu.cn, \\\{hedongliang01, zhangqi44,  lifu,  dingerrui\}@baidu.com}
}

\maketitle
\begin{abstract}

We propose a novel image retouching method by modeling the retouching process as performing a sequence of newly introduced trainable \emph{neural color operators}. The neural color operator mimics the behavior of traditional color operators and learns pixelwise color transformation while its strength is controlled by a scalar. To reflect the homomorphism property of color operators, we employ equivariant mapping and adopt an encoder-decoder structure which maps the non-linear color transformation to a much simpler transformation (i.e., translation) in a high dimensional space. The scalar strength of each neural color operator is predicted using CNN based strength predictors by analyzing global image statistics. Overall, our method is rather lightweight and offers flexible controls. Experiments and user studies on public datasets show that our method consistently achieves the best results compared with SOTA methods in both quantitative measures and visual qualities. Code is available at \href{https://github.com/amberwangyili/neurop}{\url{https://github.com/amberwangyili/neurop}}.
\end{abstract}
\section{Introduction}
\label{sec:intro}


In the digital eras, images are indispensable components for information sharing and daily communication, which makes improving the perceptual quality of images quite demanding in many scenarios.
While professional retouching software such as Adobe Photoshop and Lightroom provide various retouching operations, the effectiveness and proficiency of using them still require expertise on the part of the users.
Automation of such tedious editing works is highly desirable and has been extensively studied for decades. 

Recently, deep learning has exhibited stunning success
in many computer vision and image processing tasks~\cite{long2015fully,isola2017image}, 
which generally uses networks to learn sophisticated mappings from paired data. In the pioneering work by Bychkovsky \etal~\cite{bychkovsky2011learning}, a large scale of input and expert-retouched image pairs have been collected: the constructed MIT-Adobe FiveK dataset has enabled high-quality supervised learning and drives the mainstream towards data-driven methods. 

Among these methods, \emph{sequential image retouching}~\cite{shi2021learning,hu2018exposure,park2018distort,yan2014learning} is rather attractive.
Following human's step-by-step workflow, they model the image retouching process as a sequence of standard color operators (\ie, brightness or contrast adjustments), each of which is controlled by a scalar parameter. Besides generating a retouched image, they also produce a semantically meaningful history of retouching operators. With the retouching history on hand, the process is more interpretable, offers convenient controls for re-editing, and could be used in  applications like photo manipulation tutorial generation~\cite{Grabler2009TutorialGeneration} and image revision control~\cite{Chen2011ImageRevisionControl}.
However, due to the limited expressiveness of standard color operators, these methods require a relatively large number of operators to reproduce the complex retouching effects, which makes it less robust to accurately predict the operator parameters and hard to achieve high-fidelity retouched results (\ie, high PSNR scores) compared to other state-of-the-art retouching methods~\cite{zeng2020learning,he2020conditional,liu2021lightweight,Zhao_2021_ICCV}.


To address this issue, we introduce a novel trainable color operator, named \emph{neural color operator}. It is designed to mimic the behavior of traditional color operators, which maps a 3D RGB color to a modified color according to a scalar indicating the operator's strength. Instead of using a fixed function as in standard color operators, we allow the parameters of a neural color operator to be learned in order to model more general and complex color transformations. To reflect the \emph{homomorphism property} of color operators (\ie, adjusting exposure by $+2$, followed a second  $+3$ adjustment, is approximately equivalent to a single operator that adjusts exposure by $+5$), we employ \emph{equivariant mapping} and adopt an encoder-decoder structure, which maps the color transformation in 3D RGB space to a much simpler transformation (\ie, translation) in a high dimensional feature space where the scalar strength controls the amount of translation. 

We further propose an automatic, lightweight yet effective sequential image retouching method. We model the retouching process as applying a sequence of pixel-wise neural color operators to the input image.
The strength of each neural color operator is automatically determined by the output of a CNN based \emph{strength predictor} using global statistics of deep image features computed from downsampled intermediate adjusted images. The neural color operators and strength predictors are jointly trained in an end-to-end manner, with a carefully designed initialization scheme. 

\paragraph{\bf Contributions.} Our contributions are summarized as follows: First, we introduce neural color operator --- a novel, trainable yet interpretable color operator. It mimics the behavior of color operators whose strength is controlled by a scalar, and could 
effectively model sophisticated global color transformations. Second, based on neural color operators, we propose a lightweight and efficient method with only 28k parameters for sequential image retouching. Without any bells and whistles, experiments and user studies show that our method consistently achieves the best performance on two public datasets, including MIT-Adobe FiveK~\cite{bychkovsky2011learning} and PPR10K~\cite{Liang2021PPR10K}, compared with state-of-the-art  methods from both qualitative measures and visual qualities. Furthermore, our method inherits the nice properties of sequential image retouching, which allows flexible controls and convenient re-editing on results by intuitively adjusting the predicted scalar strengths using three sliders in real-time.

\section{Related works}
\label{sec:related}
Computational methods for automatic image retouching have kept evolving over the years --- from traditional heuristics such as
spatial filters~\cite{aubry2014fast}, to histogram-based methods~\cite{HEarici2009histogram,HEkim1997contrast,HElee2013contrast,HEstark2000adaptive,HEwang1999image}, and recent learning-based approaches. Here we give a general review on this most recent branch of researches. 

\paragraph{\bf Image-to-Image Translation based Methods.} 
These methods generally consider image retouching as a special case of image-to-image translation and use neural networks to directly generate the enhanced images from input images~\cite{ignatov2017dslr,ignatov2018wespe,kim2020pienet,chen2018deep,ni2020towards,jiang2021enlightengan,Zamir2020MIRNet,Zhao_2021_ICCV}. 
While these methods can produce visually pleasing results, they have some common limitations: they usually employ large networks and consume too much computational resources when applied to high-resolution images; besides, the use of downsampled convolutional layers may easily lead to artifacts in high frequency textures. Some methods have been proposed to alleviate these issues by
decomposition of reflectance and illumination~\cite{ying2017new,wang2019underexposed},
representative color transform~\cite{Kim_2021_ICCV}, 
or Laplacian pyramid decomposition~\cite{afifi2021learning,jie2021LPTN}. 



\paragraph{\bf Sequential Image Retouching.} This category of methods~\cite{shi2021learning,hu2018exposure,park2018distort,yan2014learning} follow the step-by-step retouching workflow of human experts, which models the process as a sequence of several color operators where the strength of each operator is controlled by a scalar. 
This modeling paradigm is quite challenging because it requires more supervision of editing sequences. Some researchers~\cite{hu2018exposure,park2018distort} resort to deep reinforcement learning algorithms which are known to be highly sensitive to their numerous hyper-parameters~\cite{henderson2018deep}. Shi \etal~\cite{shi2021learning} propose an operation-planning algorithm to generate pseudo ground-truth sequences, 
hence making the network easier to train. However, it needs extra input text as guidance. 
Overall, these methods provide an understandable editing process, and allow convenient further controls on results through intuitive adjustments of the color operators. 
However, due to the limited expressiveness of standard color operators, they require a sequence with a relatively large number of operators to reproduce complex retouching effects, which is harder to learn and leads to less satisfactory performance (\ie, lower PSNR scores). 

\paragraph{\bf Color Transformation based Methods.}
These methods usually use a low-resolution image to extract features and predict the parameters of some predefined global or local color transformation, and later apply the predicted color transformation to the original high-resolution image. Different color transformations have been used in existing works, including 
quadratic transforms~\cite{yan2016automatic,chai2020supervised,liu2020color,wang2011example}, 
local affine transforms~\cite{gharbi2017deep},
curve based transforms~\cite{bychkovsky2011learning,guo2020zero,li2020flexible,kim2020global,moran2021curl}, 
filters~\cite{deng2018aesthetic,moran2020deeplpf}, 
lookup tables~\cite{zeng2020learning,Wang_2021_ICCV}, and customized transforms~\cite{bianco2019content}. 
Compared with image-to-image translation based methods, 
these methods usually use smaller models and are more efficient. 
However, their capabilities are constrained by the predefined color transformation and might be inadequate to approximate highly non-linear color mappings between low and high quality images.

Recently, He~\etal~\cite{he2020conditional,liu2021lightweight} proposed Conditional Sequential Retouching Network (CSRNet), which models the global color transformation using a 3-layer multi-layer perceptron (MLP). The MLP is influenced through global feature modulation via a 32D conditional vector extracted from the input image through a conditional network. Their method is lightweight  and achieves nice performance. However, the 32D condition vector is hard to interpret and control. While CSRNet is named with `sequential', we prefer to regard it as a color transformation based method instead of a sequential image retouching method. This is because it does not explicitly model retouching operators or generate a meaningful retouching sequence, nor does it produce intermediate adjusted images, all of which are important features of sequential image retouching.

\paragraph{\bf Our motivation.}
We are motivated by the seemingly intractable dilemma in the context of previous retouching methods: while ``black-boxing'' the standard operators~\cite{he2020conditional,liu2021lightweight} can improve model expressiveness 
such non-interpretable parameterization leads to the lack of controllability; 
though good interpretability can be achieved by using sequential methods with standard post-processing operators~\cite{shi2021learning,hu2018exposure,park2018distort,yan2014learning}, these methods require a relatively large number of operators to reproduce the complex retouching effects but also make it less robust to accurately predict the parameters and fail to achieve satisfactory results. 

Specifically, we seek to resolve this dilemma by disentangling the complex nonlinear retouching process into a series of easier transformations modeled by our trainable neural color operators with interpretable controls,
while still maintaining high efficacy and lightweight structures. 
\section{Neural Color Operator} \label{sec:neural_operations}



\paragraph{\bf Definition.} A neural color operator (short as neural operator or neurOp) mimics the behavior of traditional global color operators. It is defined to learn a pixel-wise global color mapping $\mathcal{R}$ from an input 3D RGB color $\mathbf{p}$, with a 1D scalar $v$ indicating the operator's strength, to an output 3D RGB color $\mathbf{p'}$: 
\begin{equation} \label{equ:nop_definition}
    \mathbf{p'} = \mathcal{R}(\mathbf{p},v) .
\end{equation}
For simplicity, we restrict the scalar strength $v$ to be normalized in $[-1,1]$. 

\paragraph{\bf Color Operator Properties.} 
By carefully studying standard global color operators, such as 
\texttt{\small exposure}, \texttt{\small black clipping},  \texttt{\small vibrance} in \emph{Lightroom}, and \texttt{\small contrast}, \texttt{\small brightness} in \emph{Snapseed}, and general hue, saturation and gamma adjustments, we find those operators usually satisfy the following properties: 
\begin{itemize}
    \item 1. The effect of a color operator is controlled in a continuous way by adjusting the scalar strength. 
    \item 2. The color value should keep unchanged if strength is zero, \ie, $\mathbf{p} =\mathcal{R}(\mathbf{p},0)$. 
    \item 3. The larger the strength is, the larger the color changes. For example, adjusting brightness by $+2$ should incur more changes than $+1$ adjustment. 
    \item 4. From a mathematical aspect, the operator should approximately satisfy the \emph{homomorphism property}, \ie, $\mathcal{R}(\mathbf{p},v_1+v_2) \approx \mathcal{R}(\mathcal{R}(\mathbf{p},v_1),v_2)$. For example, adjusting exposure by $+3$, followed by a second operator that adjusts exposure by $+2$, is approximately equivalent to a single operator that adjusts exposure by $+5$. 
\end{itemize}

Recall that our goal is to mimic the behavior of traditional color operators. Hence, The above color operator properties should be considered in the design of our neural color operators. Our main observation is: since the translation function naturally satisfies those properties, \ie, for $f(x,v) = x+v$ we easily have $f(x,v_1+v_2)=f(f(x,v_1),v_2)$), 
if we could map the non-linear color transformation in the 3D RGB space 
to a simple translation function in a high-dimensional feature space, the neural color operators would also satisfy those properties.

\begin{figure*}[htpb]
\centering
\includegraphics[width=\linewidth]{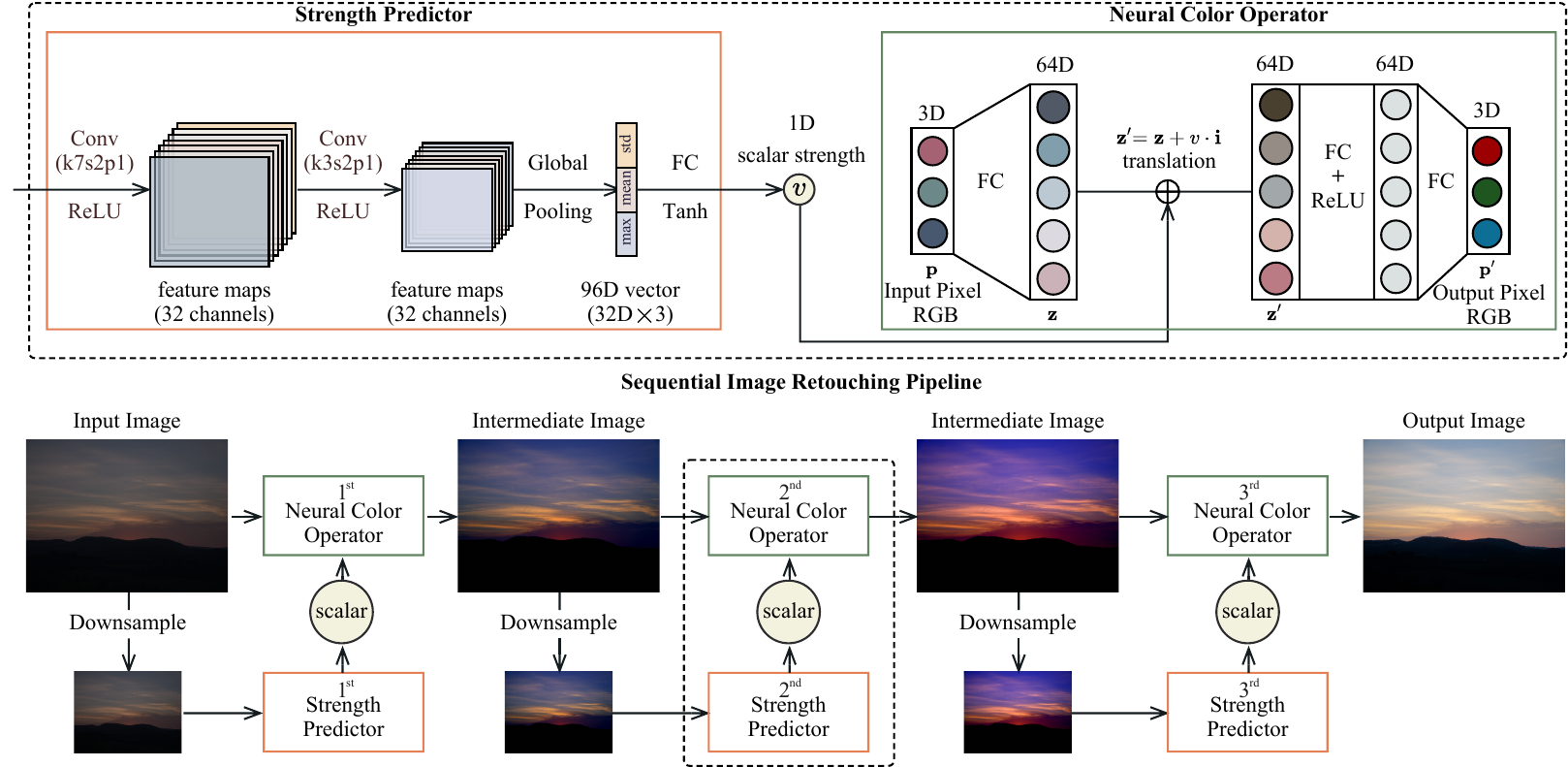}
\caption{
%
Pipeline of our approach and network structures.
Top left: structure of a strength predictor. Parameters of two convolutional layers are shared among all strength predictors (k7s2p1 denotes a kernel size $7\times7$, a stride of 2 and a padding of 1). Top right: network structure of a neural color operator. Parameters are not shared among different neural color operators.
Bottom: the overall pipeline of our method.
}
\label{fig:overview}
\end{figure*}

\paragraph{\bf Network Structure.} Based on the observation, we employ \emph{equivariant mapping}, which has been widely used in other applications like geometric deep learning~\cite{jimenez2016unsupervised,kulkarni2015deep}, for the purpose. Specifically, we adopt an encoder-decoder structure, 
as shown in Figure~\ref{fig:overview} (top right). 
The encoder $E$ transforms the input 3D RGB color $\mathbf{p}$ to a high-dimensional (\ie, 64D) feature vector $\mathbf{z}$, then we perform a simple translation in the feature space $\mathbf{z}'=\mathbf{z} + v\cdot\mathbf{i}$, where the translation direction $\mathbf{i}$ is simply set as an all-one vector in the feature space and $v$ controls the distance of translation. Finally, the decoder $D$ transforms the feature vector $\mathbf{z'}$ back to the 3D RGB space to obtain the output color $\mathbf{p'}$. In general, we define the mapping of a neural color operator $\mathcal{R}$ as:
\begin{equation}
\mathbf{p'} = \mathcal{R}(\mathbf{p},v) = D(E(\mathbf{p})+v\cdot\mathbf{i})).
\end{equation}
To maintain a lightweight structure, the encoder contains only one fully connected (FC) layer (without any hidden layers), and the decoder contains two FC layers where the first layer uses ReLU activations. 
Different from traditional encoders which usually map data to a lower dimensional latent space, our encoder maps colors to a higher dimensional feature space. We provide visualization of the 64D feature vectors in the supplemental document.

Note that our neural color operator can only approximately (but not strictly) satisfy the above color operator properties, since the decoder is not theoretically guaranteed to be the exact inverse of the encoder, i.e., $D(E(\mathbf{p})) \approx \mathbf{p}$. 

\section{Automatic Sequential Image Retouching}



In this section, we present how to develop an automatic, lightweight yet effective sequential image retouching  method based on our proposed neurOps.

\subsection{Problem Setup} \label{sec:problem_setup}
Given an input image $I\in \mathbb{R}^{H\times W \times C}$ with an arbitrary resolution, 
our goal is to generate a retouched image $I^R\in \mathbb{R}^{H\times W \times C}$ by sequentially applying $K$ pixel-wise color transformations modeled by neurOps: 
\begin{equation}
I_k = \mathcal{R}_k(I_{k-1},v_k),\quad 1\leq k \leq K, 
\end{equation}
where $I_0 = I$ is the input image, $I_k$ ($1\leq k \leq K-1$) are intermediate images, and $I^R=I_K$ is the output image. $\mathcal{R}_k$ is the $k$-th neurOp, and $v_k$ is a scalar that controls its strength. A neurOp maps each pixel's input color to an output color in a pixel-wise manner. In our experiments, we use $K=3$ neurOps for the best trade-off between model size and efficacy. $H,W$ and $C$ denote the height, width, and channel size of the image, respectively.

To make the retouching process automatic, we predict the strength of each neurOp by strength predictors: 
\begin{equation} \label{equ:strength_predictor}
v_k = \mathcal{P}_k(I^{\downarrow}_{k-1}),\quad 1\leq k \leq K, 
\end{equation}
where $\mathcal{P}_k$ is the $k$-th strength predictor, and $I^{\downarrow}_{k-1}$ is the bilinearly downsampled image from intermediate image $I_{k-1}$ by shortening the longer image edge to $256$. 
We use intermediate images instead of always using the original input image $I$ to infer the strength parameter. This is motivated by human's editing workflow: an artist also needs intermediate visual feedbacks to decide the next editing step~\cite{hu2018exposure}.  
Figure~\ref{fig:overview} (bottom) illustrates the overall pipeline of our approach. Note that we do not share parameters between the 3 neurOps.

\subsection{Strength Predictor}

A strength predictor is used to predict the scalar strength used in a neurOp. Global image information is required for such prediction, i.e., an exposure adjustment needs the average intensity of the image. Hence, we use a downsampled intermediate edited image from the previous editing step as the input to the strength predictor. 

The strength predictor contains 4 layers: two convolutional layers, a global pooling layer, and a FC layer. First, we use the first two convolutional layers to extract 32-channel  feature maps. Then, we apply global pooling to obtain a 96D vector. Specifically, we employ three different pooling functions to compute channel-wise maximum, average, and standard deviation, respectively, and concatenate the three 32D vectors together. Finally, the FC layer, which acts as a predicting head, maps the 96D vector to a scalar strength in $[-1,1]$. 
The top left of Figure~\ref{fig:overview} shows the network structure of a strength predictor.


To reduce the overall size of our network, we force the parameters of the two convolutional layers 
to be shared among all 3 strength predictors. The parameters of the predicting head (\ie, the last FC layer) are not shared. 

Our strength predictor is inspired by the conditional network in CSRNet~\cite{liu2021lightweight,he2020conditional} in order to maintain a lightweight structure. It is improved in several ways. First, we take intermediate images instead of the original image as input to better reflect human's editing workflow. Second, their conditional network outputs a 32D vector while our strength predictor generates a scalar. Third, we use three pooling functions instead of only one average pooling function which are demonstrated to be more helpful.

\subsection{Loss Function and Training}

We jointly train our neurOps and strength predictors by optimizing a predefined loss function in an end-to-end fashion. In practice, we find that a carefully designed initialization for the neurOps is also rather helpful. 

\paragraph{\bf Loss Function.} 
For a single training image $I$, denoting its ground truth retouched image as $I^{GT}$ and the predicted retouched image as $I^R$, respectively, the loss $\mathcal{L}$ is defined as the weighted sum of a reconstruction loss $\mathcal{L}_{r}$, a total variation (TV) loss $\mathcal{L}_{tv}$, and a color loss $\mathcal{L}_{c}$: 
\begin{equation} \label{equ:loss_total}
\mathcal{L} = \mathcal{L}_{r} +  \lambda_1\mathcal{L}_{tv}+\lambda_2\mathcal{L}_{c},
\end{equation}
 where $\lambda_1$ and $\lambda_2$ are two balancing weights empirically set as $\lambda_1=\lambda_2=0.1$. The reconstruction loss measures L1 difference between the predicted image and the ground truth:
\begin{equation} \label{equ:loss_l1}
\mathcal{L}_{r} = \frac{1}{CHW}\|I^R -I^{GT}\|_1.
\end{equation}
A TV loss~\cite{aly2005image} is included to encourage spatially coherent predicted images:
\begin{equation} \label{equ:loss_tv}
\mathcal{L}_{tv}= \frac{1}{CHW}\|\nabla I^R  \|_2,
\end{equation} 
where $\nabla(\cdot)$ denotes the gradient operator. 
Besides, we include a color loss \cite{liu2020color,wang2019underexposed}, which regards RGB colors as 3D vectors and measures their angular differences. Specifically, the color loss is defined as:
\begin{equation}
    \mathcal{L}_c = 1-\frac{1}{HW}\angle(I^R,I^{GT}),
\end{equation}
where $\angle(\cdot)$ is an operator that computes the pixel-wise average of the cosine of the angular differences. The overall loss is computed as the sum of losses (Equation~\ref{equ:loss_total}) over all training images. 

\paragraph{\bf Initialization Scheme for Neural Color Operators.}
Instead of initializing the neurOps from scratch, we find 
that initializing them with standard color operators is a better choice. By analyzing the retouching histories in the MIT-Adobe FiveK dataset~\cite{bychkovsky2011learning}, we picked up the 3 
most commonly used standard operators in the dataset, which are 
\texttt{\small black clipping}, \texttt{\small exposure}, and 
\texttt{\small vibrance} in \emph{Lightroom}. For initialization, we intend to let our 3 neurOps reproduce the above 3 standard operators, respectively. 

Since the formulas of those Lightroom operators are not publicly available, we have selected a small number of images (\ie, 500 images) from the training set of MIT-Adobe FiveK, and manually retouched those images using those standard operators. Specifically, for each standard operator, for each input image $I$,  we manually generate $M$ (\ie, $M=40$) retouched images (\ie, $\{I_m\}$, $1\leq m \leq M$) with uniformly distributed levels of strengths. The strength used for each retouched image $I_m$ is denoted as $v_m$. 

Then, we enforce each neurOp $\mathcal{R}$ to reproduce the manually retouched images from each standard operator as much as possible. To do so, we optimize each neurOp by minimizing the sum of a unary loss and a pairwise loss over all selected images. 
The unary loss $\mathcal{L}_{1}$ enforces that the color keeps unchanged when the strength is zero: 
\begin{equation}
\mathcal{L}_{1} = \frac{1}{M}\sum_{m}{\|\mathcal{R}(I_{m},0) - I_{m} \|_1}.
\end{equation}
The pairwise loss $\mathcal{L}_{2}$ enforces that the neurOp 
behaves the same as the given standard operator for specific strength values: 
\begin{equation}
\mathcal{L}_{2} = \frac{1}{M(M-1)}\sum_{m\neq 
n}{\|\mathcal{R}(I_{m},v_n-v_m) - I_{n} \|_1}.
\end{equation}

For optimization, we use the Adam optimizer with an initial 
learning rate of $5e^{-5}$ , $\beta_1= 0.9$, $\beta_2 = 0.99$, and 
a mini-batch size of 1, and run 100,000 iterations. 

\paragraph{\bf Joint Training.} After initializing neurOps 
using the scheme described above, and initializing the strength 
predictors with random parameters, we jointly train our neurOps and strength predictors in an end-to-end fashion towards 
minimizing the overall loss function 
(Equation~\ref{equ:loss_total}). We also use the Adam optimizer 
with the same configurations as in the initialization scheme, and 
run 600,000 iterations. For data augmentation, we randomly crop 
images and then rotate them by multiples of 90 degrees.


\section{Experiments}
\label{sec:exp}


\paragraph{\bf Datasets.} The MIT-Adobe FiveK dataset~\cite{bychkovsky2011learning} has been widely used for global image retouching and enhancement tasks. It consists of 5000 images in RAW format and an Adobe Lightroom Catalog that contains the adjusted rendition setting by five experts. We follow the common practice and use rendition version created by expert C as the ground truth. There are two public variations in terms of input rendition setting. The first variation is provided by Hu \etal~\cite{hu2018exposure}, which choose the input \texttt{\small with Daylight WhiteBalance minus 1.5} and export them to 16 bits ProPhotoRGB TIFF images. We  refer to it as \emph{MIT-Adobe-5K-Dark}. The second variation is provided by Hwang \etal~\cite{hwang2012context} which adopt the input
\texttt{\small As Shot Zeroed} and export them to 8 bits sRGB JPEG images. We refer to it as \emph{MIT-Adobe-5K-Lite}. 

PPR10K~\cite{Liang2021PPR10K} is a recently released image retouching dataset. It contains more than 11,000 high-quality portrait images. For each image, a human region mask is given and three retouched images by three experts are provided as ground truth. We refer to the three retouched variations as \emph{PPR10K-a}, \emph{PPR10K-b}, and \emph{PPR10K-c}, respectively. 

We train and evaluate our method on all 5 variations of both datasets. We follow the same split of training and testing sets as in previous works. For the MIT-Adobe FiveK dataset, all images are resized by shortening the longer edge to 500. For the PPR10K dataset, we use image resolution of 360p.  
All the experiments are performed on a PC with an NVIDIA RTX2080 Ti GPU. It takes about 2 hours for initialization, and takes about 9 and 15 hours for training on MIT-Adobe FiveK and PPR10K, respectively.



\subsection{Comparison and Results}
\label{subsec:res}
\paragraph{\bf Comparison.} We compare our method with a series of state-of-the-art methods, including 
White-Box~\cite{hu2018exposure}, Distort-and-Recover~\cite{park2018distort}, HDRNet~\cite{gharbi2017deep}, DUPE~\cite{wang2019underexposed}, MIRNet~\cite{Zamir2020MIRNet}, Pix2Pix~\cite{isola2017image},
3D-LUT~\cite{zeng2020learning}, CSRNet~\cite{he2020conditional,liu2021lightweight} on \emph{MIT-Adobe-5K-Dark}, 
and DeepLPF~\cite{moran2020deeplpf}, IRN~\cite{Zhao_2021_ICCV} on \emph{MIT-Adobe-5K-Lite}, 
and CSRNet, 3D-LUT, 3D-LUT+HRP~\cite{Liang2021PPR10K} on \emph{PPR10K}. 

For the PPR10K dataset, since each image is associated with a human region mask, we also provide a variant of our method that additionally considers human region priority (HRP)~\cite{Liang2021PPR10K} in training, which we refer to as NeurOp+HRP. This is done by slightly modifying the L1 loss in Equation~\ref{equ:loss_l1} to be a weighted one, \ie, pixels inside human region have larger weights (e.g., 5) while other pixels have smaller weights (e.g., 1). 

As shown in Table~\ref{tab:sota}, our method consistently achieves the best performance on all dataset variations, demonstrating the robustness of our method. In contrast, while CSRNet performs relatively well on MIT-Adobe-5K-Dark, it performs less satisfactory on PPR10K, \ie, their PSNR is lower than ours by 2db on PPR10K-a, which is consistent with the benchmark carried out in~\cite{Liang2021PPR10K}. Furthermore, our method is the most lightweight one (\ie, only 28k parameters). 

\begin{table}[t]
\centering
\caption{Quantitative comparison with state-of-the-art methods. 
			The best results are boldface and the second best ones are underlined.
%
} 

\resizebox{\textwidth}{!}{ 
\begin{tabular}{clcccl}
			\toprule
			Dataset & Method & PSNR$\uparrow$ & SSIM$\uparrow$ 
			& $\Delta E^*_{ab}$$\downarrow$
			& \#params\\
			\midrule
		\multirow{9}{*}{\shortstack{MIT\\-Adobe\\-5K-Dark}}	
		& White-Box \cite{hu2018exposure}  & 18.59 & 0.797 & 17.42  & 8,561,762\\
					\noalign{\smallskip}
			& Dis.\& Rec. \cite{park2018distort}  & 19.54 & 0.800 & 15.44 &  259,263,320\\
						\noalign{\smallskip}
			& HDRNet \cite{gharbi2017deep}  & 22.65 & 0.880 & 11.83  & 482,080\\
						\noalign{\smallskip}
			& DUPE \cite{wang2019underexposed} & 20.22 & 0.829 & 16.63 & 998,816\\
						\noalign{\smallskip}
			& MIRNet \cite{Zamir2020MIRNet}  & 19.37 & 0.806  & 16.51 & 31,787,419\\
						\noalign{\smallskip}
			& Pix2Pix \cite{isola2017image}  & 21.41 & 0.749  & 13.26 & 11,383,427\\
						\noalign{\smallskip}
			& 3D-LUT \cite{zeng2020learning}  & 23.12 & 0.874 & 11.26  & 593,516\\
						\noalign{\smallskip}
			& CSRNet \cite{he2020conditional,liu2021lightweight}  & \underline{23.86} & \underline{0.897} & \underline{10.57} & \underline{36,489}\\
						\noalign{\smallskip}
			& NeurOp (ours) & \textbf{24.32} & \textbf{0.907}  & \textbf{10.10} & \textbf{28,108}\\
\noalign{\smallskip}
\midrule	
\noalign{\smallskip}
		\multirow{3}{*}{\shortstack{MIT-\\Adobe\\-5K-Lite}}	
			& DeepLPF \cite{moran2020deeplpf} & 23.63 & 0.875 & 10.55  & \underline{1,769,347}\\
						\noalign{\smallskip}
			& IRN \cite{Zhao_2021_ICCV} & \underline{24.27} &  \underline{0.900} & \underline{10.16} & 11,650,752\\			\noalign{\smallskip}
			& NeurOp (ours)  & \textbf{25.09} & \textbf{0.911} & \textbf{9.93} & \textbf{28,108}\\
			\noalign{\smallskip}
			\bottomrule
		\end{tabular}
\quad		
\begin{tabular}{clccc}
			\toprule
			Dataset & Method & PSNR$\uparrow$&SSIM$\uparrow$ 
			& $\Delta E^*_{ab}$$\downarrow$\\
			\midrule
		\multirow{5}{*}{\shortstack{PPR\\10K-a}}	
		    & CSRNet \cite{he2020conditional,liu2021lightweight} & 24.24 & 0.937 & 9.75 \\
  			& 3D-LUT \cite{zeng2020learning}  & 25.64  & - & -\\
			& 3D-LUT+HRP \cite{Liang2021PPR10K}  & 25.99 & 0.952 & 8.95\\
			& NeurOp (ours) & \underline{26.32} & \underline{0.953} & \underline{8.81}\\
			& NeurOp+HRP (ours) & \textbf{26.46} & \textbf{0.955} & \textbf{8.80}\\
\midrule	
		\multirow{5}{*}{\shortstack{PPR\\10K-b}}	
		    & CSRNet \cite{he2020conditional,liu2021lightweight} & 23.93 & 0.938 & 9.83\\
			& 3D-LUT \cite{zeng2020learning}  & 24.70 &- & -  \\
			& 3D-LUT+HRP \cite{Liang2021PPR10K}  & 25.06 & 0.945 & 9.36\\
			& NeurOp (ours) & \underline{25.45} & \underline{0.946}& \underline{9.21}\\
			& NeurOp+HRP (ours) & \textbf{25.82} & \textbf{0.951} & \textbf{8.97}\\
\midrule	
		\multirow{5}{*}{\shortstack{PPR\\10K-c}}
		    & CSRNet \cite{he2020conditional,liu2021lightweight} & 24.35 & 0.929 & 9.92 \\
			& 3D-LUT \cite{zeng2020learning}  & 25.18 &- & -  \\
			& 3D-LUT+HRP \cite{Liang2021PPR10K}  & 25.46 & 0.939 & 9.43\\
			& NeurOp (ours) & \underline{26.02}  & \underline{0.946} & \underline{8.94}\\
			& NeurOp+HRP (ours) & \textbf{26.23} & \textbf{0.947} & \textbf{8.86}\\
			\bottomrule
		\end{tabular}
} 
\label{tab:sota}
\end{table}

\begin{figure}[t]
	\centering

\resizebox{\textwidth}{!}{ 
	\begin{tabular}{cccccc}
	   \multicolumn{6}{c}{\scalebox{0.8}{\textbf{Dataset: MIT-Adobe-5K-Dark}}} \\
		\includegraphics[width=0.16\linewidth]{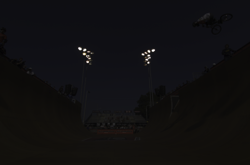}&
		\includegraphics[width=0.16\linewidth]{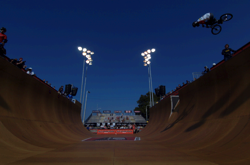}&
		\includegraphics[width=0.16\linewidth]{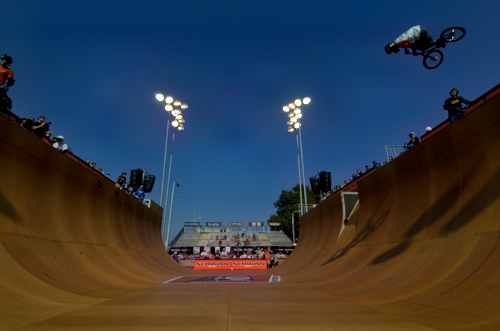}&
		\includegraphics[width=0.16\linewidth]{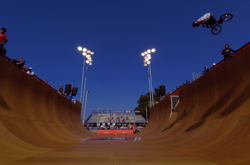}&
		\includegraphics[width=0.16\linewidth]{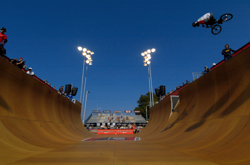}&
		\includegraphics[width=0.16\linewidth]{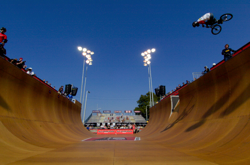}\\
		
		\includegraphics[width=0.16\linewidth]{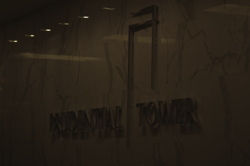}&
		\includegraphics[width=0.16\linewidth]{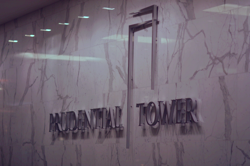}&
		\includegraphics[width=0.16\linewidth]{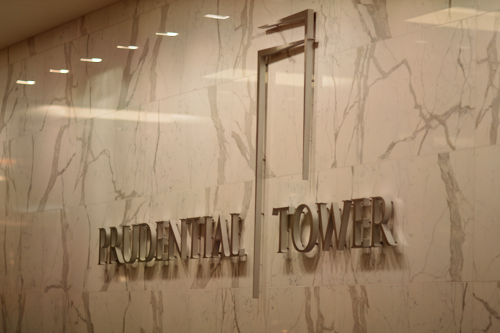}&
		\includegraphics[width=0.16\linewidth]{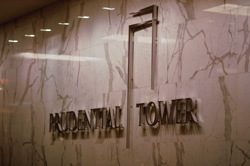}&
		\includegraphics[width=0.16\linewidth]{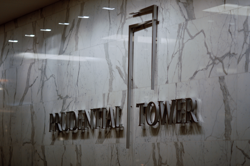}&
		\includegraphics[width=0.16\linewidth]{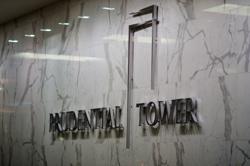}\\
		\scalebox{0.8}{Input}
		& \scalebox{0.8}{3D-LUT~\cite{zeng2020learning}}
		& \scalebox{0.8}{HDRNet~\cite{gharbi2017deep}}
		& \scalebox{0.8}{CSRNet~\cite{he2020conditional,liu2021lightweight}}
		& \scalebox{0.8}{NeurOp (ours)} 
		& \scalebox{0.8}{GT}
	\end{tabular}}

\resizebox{\textwidth}{!}{ 
	\begin{tabular}{ccccc}
	\multicolumn{5}{c}{\scalebox{0.8}{\textbf{Dataset: MIT-Adobe-5K-Lite}}} \\
		\includegraphics[width=0.19\linewidth]{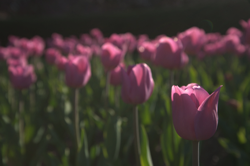}&
		\includegraphics[width=0.19\linewidth]{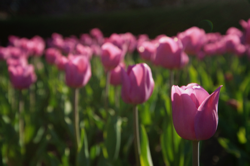}&					
		\includegraphics[width=0.19\linewidth]{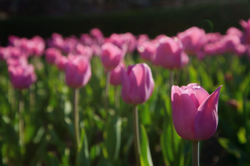}&				
		\includegraphics[width=0.19\linewidth]{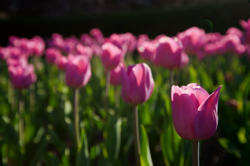}&
		\includegraphics[width=0.19\linewidth]{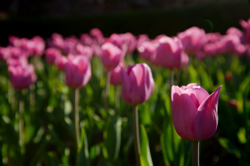}\\
		\includegraphics[width=0.19\linewidth]{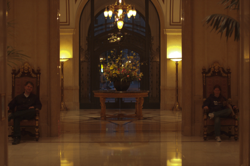}&
		\includegraphics[width=0.19\linewidth]{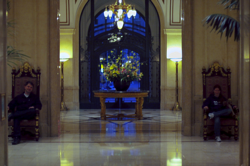}&
		\includegraphics[width=0.19\linewidth]{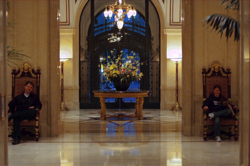}&
		\includegraphics[width=0.19\linewidth]{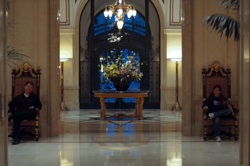}&
		\includegraphics[width=0.19\linewidth]{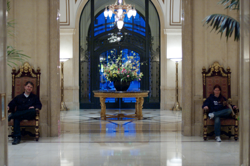}\\
		
		\scalebox{0.8}{Input}
		& \scalebox{0.8}{DeepLPF~\cite{moran2020deeplpf}}
		& \scalebox{0.8}{IRN~\cite{Zhao_2021_ICCV}}
		& \scalebox{0.8}{NeurOp (ours)}
		& \scalebox{0.8}{GT}
	\end{tabular}}
	\resizebox{\textwidth}{!}{ 
	\begin{tabular}{cccccc}
	\multicolumn{6}{c}{\scalebox{0.8}{\textbf{Dataset: PPR10K-a}}} \\
		\includegraphics[width=0.16\linewidth]{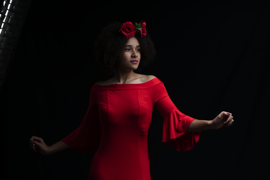}&
		\includegraphics[width=0.16\linewidth]{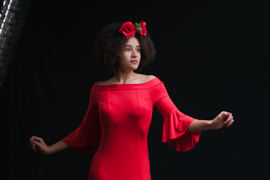}&
		\includegraphics[width=0.16\linewidth]{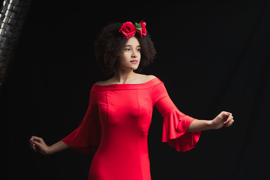}&
		\includegraphics[width=0.16\linewidth]{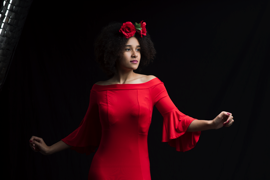}&
		\includegraphics[width=0.16\linewidth]{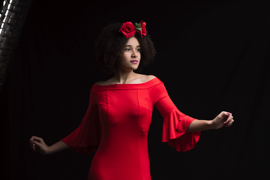}&
		\includegraphics[width=0.16\linewidth]{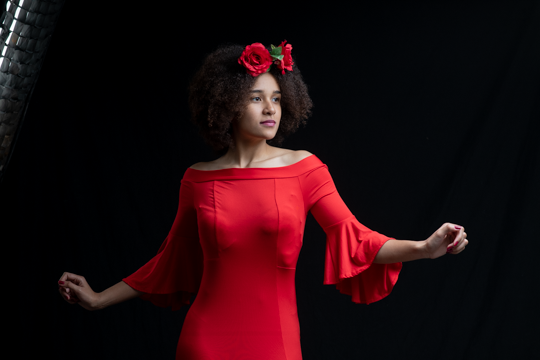}\\
		\includegraphics[width=0.16\linewidth]{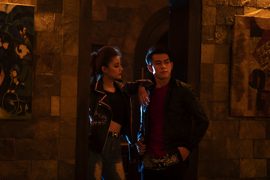}&
		\includegraphics[width=0.16\linewidth]{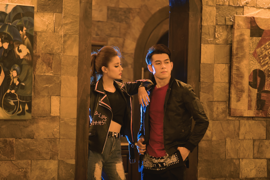}&
		\includegraphics[width=0.16\linewidth]{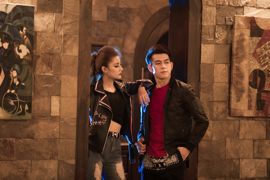}&
		\includegraphics[width=0.16\linewidth]{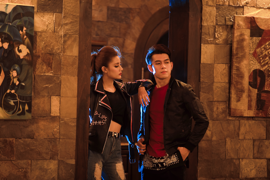}&
		\includegraphics[width=0.16\linewidth]{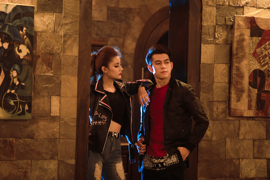}&
		\includegraphics[width=0.16\linewidth]{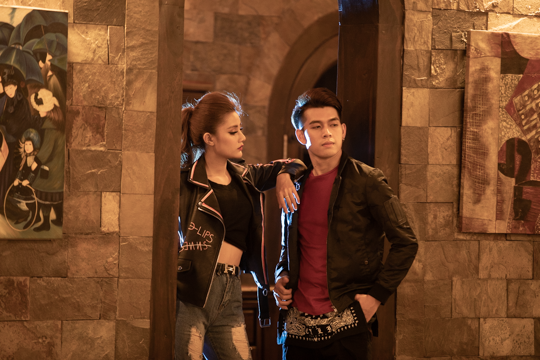}\\
		\scalebox{0.8}{Input}
		& \scalebox{0.8}{CSRNet}
		& \scalebox{0.8}{3D-LUT+HRP}
		& \scalebox{0.8}{NeurOp}
		& \scalebox{0.8}{NeurOp+HRP} 
		& \scalebox{0.8}{GT}
	\end{tabular}}	
	
	\caption{Visual comparison with state-of-the-art methods. 
	}
	\label{fig:main_mit5k_lite}
\end{figure}


Figure~\ref{fig:main_mit5k_lite} shows visual comparisons. 
3D-LUT sometimes generates color banding artifacts due to the use of color space interpolation. Other methods easily lead to color shifting problems especially when the input image has a very low exposure or has a very different temperature. Generally, the results of our method have fewer artifacts and are closer to the ground truth. More visual comparisons are given in the supplemental document.




\begin{figure}[t]
\centering
\begin{tabular}{cc}
 \raisebox{-78pt}{\includegraphics[width=0.35\columnwidth]{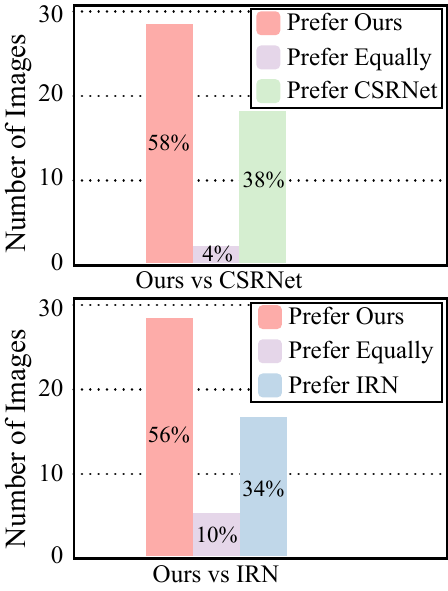}}  &  
\begin{tabular}{cccc}
\raisebox{16pt}{\scalebox{0.8}{\rotatebox[origin=c]{90}{1st neurOp}}}
&\includegraphics[width=0.18\linewidth]{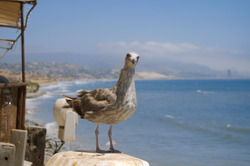}
&\includegraphics[width=0.18\linewidth]{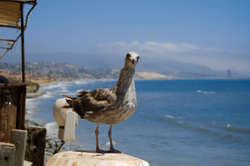}
&\includegraphics[width=0.18\linewidth]{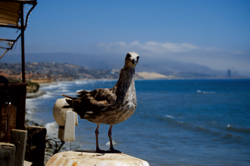}
\\
\raisebox{16pt}{\scalebox{0.8}{\rotatebox[origin=c]{90}{2nd neurOp}}}
&\includegraphics[width=0.18\linewidth]{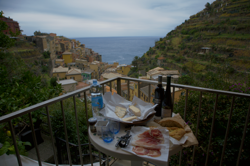}
&\includegraphics[width=0.18\linewidth]{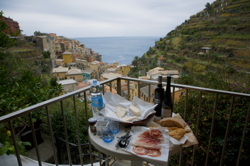}
&\includegraphics[width=0.18\linewidth]{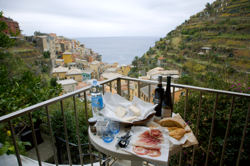}
\\
\raisebox{16pt}{\scalebox{0.8}{\rotatebox[origin=c]{90}{3rd neurOp}}}
&\includegraphics[width=0.18\linewidth]{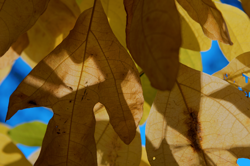}
&\includegraphics[width=0.18\linewidth]{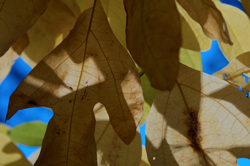}
&\includegraphics[width=0.18\linewidth]{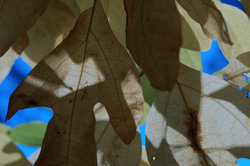} \\
& 
\multicolumn{3}{c}{
\begin{tikzpicture}
		\draw[->] (-0.3, 0) -- (1.5, 0);
		\draw[<-] (-4.5, 0) -- (-2.7, 0);
		\draw (-1.5, 0) node {\scalebox{0.8}{predicted image}};
		\draw (0.6, 0.2) node {\scalebox{0.8}{increase strength}};
		\draw (-3.6, 0.2) node {\scalebox{0.8}{decrease strength}};
\end{tikzpicture}	
}
\end{tabular}     
    \\
    (a) Results of user study  & (b) Examples of controllability
\end{tabular}
\caption{ User study and controllability.
(a) Results of user study. 
(b) Examples of controllability. The middle column shows our automatically retouched images. The left and right columns show the results by further adjusting strengths of specific neurOps.
}
\label{fig:userstudy_and_control}
\end{figure}

\paragraph{\bf User Study.} We have conducted a user study to evaluate the subjective visual quality of our method on both dataset variations of MIT-Adobe-5K. For MIT-Adobe-5K-Dark, we compare with  CSRNet~\cite{he2020conditional,liu2021lightweight}, while for MIT-Adobe-5K-Lite, we choose IRN~\cite{Zhao_2021_ICCV} for comparison, since the two methods achieve the second best performance on the two variations, respectively. 
For each dataset variation, we randomly select 50 images from the testing set and invite 10 participants (totally 20 participants are invited). For each selected image, for each participant, three retouched images are displayed: the ground truth image, and the two images generated by our method and the competing method (the latter two are displayed in random order). 
The participant is asked to vote 
which one of the latter two is visually more pleasing and more similar to the ground truth. 
Figure~\ref{fig:userstudy_and_control} (a) shows the results of user study, which suggest that our retouched images are visually more appreciated than those of competing approaches.

\paragraph{\bf Controllability.} While our method could generate the retouched images in a fully automatically way, 
we still allow users
to intuitively adjust the results by changing the predicted scalar strengths using three sliders in real-time.
Figure~\ref{fig:userstudy_and_control} (b) shows some examples. 
We find that adjusting the strengths could result in meaningful edits. For example, increasing the strength of the first neurOp amplifies the shadow and adds more contrast. Increasing the second tends to give more brightness to the middle tone while preserving the highlight proportion unchanged. Adjusting the third makes the image look cooler or warmer. 
More examples of controllability as well as the intermediate images are provided in the supplemental document. 

Existing methods such as CSRNet~\cite{he2020conditional,liu2021lightweight} achieve controlling through a linear interpolation between the retouched image and the original image by adjusting the interpolation weight. However, such controllability is rather limited. First, linear interpolation cannot faithfully reproduce sophisticated and highly non-linear color transformations. Second, only one degree of freedom is usually not enough for detailed color adjustments.

\paragraph{\bf Timing and Memory Consumption.} 
Our method runs in real-time. It takes 4 ms for a 500 × 333 image or 19 ms for an image with 1M pixels, which is slightly slower than CSRNet and 3D-LUT. However, it is worthy since our method has smaller model size (\ie, only 28k parameters), achieves better performance (\ie, higher PSNR scores), and inherits advantages of sequential image retouching which they do not possess. 

For memory consumption, our strength predictor takes less than 5M, our neroOp takes less than 1K per pixel. Overall, by processing pixels in batches (\ie, 4096 pixels a batch), our method takes less than 10M memory for input images with arbitrary resolutions (storage for input/output images not included). 

\paragraph{\bf Benefits of Our Method.} Due to the nice properties of neural color operators, the advantages of our method are multi-fold. First, it is rather lightweight and runs in real-time, hence, it could be easily deployed in mobile devices. Second, it inherits the advantages of sequential image retouching methods: providing an understandable editing history and offering convenient and flexible controls. Last but not least, besides the above advantages, our method still consistently achieves the best performance compared to state-of-the-art methods on two public datasets in both quantitative measures and visual qualities, as demonstrated by experiments and user studies.

\begin{table}[t]
\centering
\caption{ Ablation study for neural color operators.  
(a) Different initialization schemes. 
(b) Initialization with different ordering of standard operators. $\protect\overrightarrow{\text{vbe}}$ denotes using the order of \texttt{\small vibrance}, \texttt{\small black clipping}, and \texttt{\small exposure} for initialization. 
(c) Different number of neurOps. 
(d) Our method (final model).
} 
\label{tab:abla_nop}
\scalebox{0.8}{
\begin{tabular}{c|c|c|c|c|c|c|c|c|c|c|c}
\toprule
\multirow{2}{*}{Configs} & \multicolumn{2}{c}{(a) Initialization} & \multicolumn{5}{|c|}{(b) Order} & \multicolumn{3}{c|}{(c) Number} & 
(d) Ours \\
 & random & standard fix 
 & $\overrightarrow{\text{vbe}}$ & $\overrightarrow{\text{veb}}$ & $\overrightarrow{\text{evb}}$ & $\overrightarrow{\text{ebv}}$ & $\overrightarrow{\text{bve}}$ 
 & $K$=1 & $K$=2 & $K$=4 
 & finetune, $\overrightarrow{\text{bev}}$, $K$=3 \\
 PSNR$\uparrow$ & 22.61 & 23.78 & 23.93 & 24.17 & 24.11 & 24.12 & 24.23 & 21.97 & 23.5 & 24.17 & 24.32 \\ 
\bottomrule
\end{tabular}
}
\end{table}

\begin{table}[t]
\centering
%
%
\caption{ 
Ablation study for strength predictors and loss function. (a) Different pooling configurations in strength predictors. (b) Other design components in strength predictors. 
(c) Different configurations for the loss function. 
(d) Our method (final model).
%
} 
\label{tab:abla_pred}
\scalebox{0.8}{
\begin{tabular}{c|c|c|c|c|c|c|c|c|c}
\toprule
\multirow{2}{*}{Configs} & \multicolumn{3}{c}{(a) Pooling Layer} & \multicolumn{2}{|c|}{(b) Strength Predictor} & \multicolumn{3}{c|}{(c) Loss Function} & 
\multirow{2}{*}{(d) Ours} \\
 & aver & aver+max  & aver+std & ori-img-input & non-share-layer
& $\mathcal{L}_{r}$ & $\mathcal{L}_{r}$ + $\mathcal{L}_{c}$ & $\mathcal{L}_{r}$ + $\mathcal{L}_{tv}$ 
& \\
PSNR$\uparrow$ & 23.96 &24.27 &24.11 & 23.91 & 24.35 & 24.22 & 24.30 & 24.26 & 24.32 \\ 
\#Params & 27,916 & 28,012 & 28,012 & 28,108 & 56,076 & 28,108 & 28,108 & 28,108 & 28,108  \\
\bottomrule
\end{tabular}
}
\end{table}

\subsection{Ablation Study}
\label{subsec:abla}

We conduct a series of ablation studies to justify
our training strategy as well as the design choices of our neurOps and strength predictors. Except for the ablated parts, we retrain the ablated models adopting the same setting as our final models. We make quantitative comparisons using average PSNR achieved on the testing images of MIT-Adobe-5K-Dark as it is generally more challenging. 


\paragraph{\bf Neural Color Operators.} 
We first verify the effectiveness of the initialization strategy for neurOps. Recall that we first initialize all neurOps with standard operators and then finetune their parameters in a later joint training step. We compare our choice with two alternatives: random initialization from scratch, initialization with standard operators without allowing further finetuning (\ie, in later steps, only strength predictors are trained while neurOps keep fixed). The performance is given in Table~\ref{tab:abla_nop} (a) and (d). The results verify that initialization using standard operators is clearly superior to random initialization (PSNR: 24.32db vs 22.61db). Besides, using trainable neurOps is also shown to be a better choice than using fixed functional standard operators (PSNR: 23.78db), due to better model expressiveness. 

Recall that we choose the three most commonly used standard operators to initialize our neurOps. We then test whether the order of standard operators matters. As shown in Table~\ref{tab:abla_nop} (b) and (d), we test all 6 permutations of different ordering, and find that initializing the first, second and third neural operators with \texttt{\small black clipping}, \texttt{\small exposure} and \texttt{\small vibrance}, respectively, is the best choice. 



We further conduct experiments to find out how many neurOps are suitable. We test four choices: $K$ = 1, 2, 3 and 4. For $K$ = 1, we use \texttt{\small black clipping} for initialization. For $K$ = 2, we use \texttt{\small black clipping} and \texttt{\small exposure} for initialization. For $K$ = 4, we use another standard operator \texttt{\small highlight recovery} to initialize the 4th neurOp. We could find that $K=3$ achieves the best performance, as shown in Table~\ref{tab:abla_nop} (c-d). This is possibly due to that fewer neural operators (\ie, $K$ = 1 or 2) 
lead to insufficient expressiveness while a larger number of neural operators (\ie, $K$ = 4) result in harder training of strength predictors due to longer sequence. Overall, $K$ = 3 is the best choice. 


\paragraph{\bf Strength Predictors.}
In the global pooling layer of a strength predictor, recall that we have combined three pooling functions that compute average, maximum, and standard deviation, respectively. We test cases when a part of or all pooling functions are used, as shown in Table~\ref{tab:abla_pred} (a) and (d). Notice that each pooling function is useful, and combining all of them achieves the best performance.

We also verify other design choices of strength predictors in Table~\ref{tab:abla_pred} (b). Recall that we feed intermediate images from previous editing steps as the input to strength predictors. An alternative choice would be always feeding the original input image, however, that would result in a drop in performance, \ie, PSNR: 24.32db$\rightarrow$23.91db. It verifies that following human's editing workflow to use intermediate visual feedbacks is a better choice. Besides, recall that we force all strength predictors to share parameters for convolutional layers. We also test the case when parameters are not forced sharing. 
Unsurprisingly, the performance slightly gains, \ie, increasing PSNR by 0.03db. However, it doubles the network size (28k $\rightarrow$ 56k). Sharing parameters for all strength predictors leads to a good trade-off between maintaining high performance and lightweight structures. 



\paragraph{\bf Loss Function.} Our loss function (Equation~\ref{equ:loss_total}) includes a reconstruction loss term, a TV loss term, and a color loss term. In Table~\ref{tab:abla_pred} (c), we evaluate the choice of terms in the loss function. The results verify that combining all terms produces the best performance. 



\section{Conclusion and Limitations}

In this paper, we have proposed a lightweight sequential image retouching method. The core of our method is the newly introduced neural color operator. It mimics the behavior of traditional color operators and learns complex pixel-wise color transformation whose strength is controlled by a scalar. We have also designed CNN based strength predictors to automatically infer the scalar strengths . The neural color operators and strength predictors are trained in an end-to-end manner together with a carefully designed initialization scheme. 

Extensive experiments show that our model achieve the state-of-the-art performance on public datasets with only 28k parameters and provide more convenient parametric controls compared with previous competitive works.

\paragraph{\bf Limitations.} Our method has several limitations and could be further improved in the future. First, our method is limited to global color retouching. A possible way to handle local effects is extending strength predictors to infer strength maps instead of scalar strengths. Second, we only support pixel-wise color editing. It would be an interesting topic to extend neural color operators to handle spatial filtering. Third, from a theoretical aspect, our encoder-decoder structure for neural color operators cannot guarantee accurate homomorphism properties. It is worthwhile to investigate other network structures with better theoretical properties, such as invertible networks. 
As for future works, we are also interested in applying the idea of neural color operators to other related applications such as image editing.

\paragraph{\bf Acknowledgements.}
This work is supported by the National Natural Science Foundation of China (Project Number: 61932003).

%
%
\bibliographystyle{splncs04}
\bibliography{main}
\end{document}